\documentclass[12pt,onecolumn]{IEEEtran}
\usepackage{subcaption}
\IEEEoverridecommandlockouts
\usepackage{cite}
\usepackage{amsmath,amssymb,amsfonts}
\usepackage{algorithmic}
\usepackage{graphicx}
\usepackage{multirow}
\usepackage{textcomp}
\usepackage{xcolor}
\usepackage[numbers,sort&compress]{natbib}
\usepackage[colorlinks=true,linkcolor=blue,citecolor=blue,urlcolor=blue]{hyperref}
\def\BibTeX{{\rm B\kern-.05em{\sc i\kern-.025em b}\kern-.08em
    T\kern-.1667em\lower.7ex\hbox{E}\kern-.125emX}}

\newcommand{\Remark}[1]{{\color{red}}}

\begin{document}
\title{Modeling Electric Vehicle Car-Following Behavior: Classical vs Machine Learning Approach}
% \title{A Comparative Analysis of Car-Following Behavior of Electric Vehicles: Classical vs Machine Learning Approach}
%Comparative Modeling and Calibration of EV Car-Following Behavior Using Classical and Machine Learning Approaches%

\author{\IEEEauthorblockN{Md. Shihab Uddin\IEEEauthorrefmark{1},
Md Nazmus Shakib\IEEEauthorrefmark{1}, and
Rahul Bhadani\IEEEauthorrefmark{1}}\\
\IEEEauthorblockA{\IEEEauthorrefmark{1}Electrical and Computer Engineering,
The University of Alabama in Huntsville, Huntsville, AL, USA\\
Emails: mu0018@uah.edu, nazmus.shakib@uah.edu, rahul.bhadani@uah.edu}
}

\maketitle

\begin{abstract}
The increasing adoption of electric vehicles (EVs) necessitates an understanding of their driving behavior to enhance traffic safety and develop smart driving systems. This study compares classical and machine learning models for EV car-following behavior. Classical models include the Intelligent Driver Model (IDM), Optimum Velocity Model (OVM), Optimal Velocity Relative Velocity (OVRV), and a simplified CACC model, while the machine learning approach employs a Random Forest Regressor. Using a real-world dataset of an EV following an internal combustion engine (ICE) vehicle under varied driving conditions, we calibrated classical model parameters by minimizing the RMSE between predictions and real data. The Random Forest model predicts acceleration using spacing, speed, and gap type as inputs. Results demonstrate the Random Forest’s superior accuracy, achieving RMSEs of 0.0046 (medium gap), 0.0016 (long gap), and 0.0025 (extra-long gap). Among physics-based models, CACC performed best, with an RMSE of 2.67 for long gaps. These findings highlight the machine learning model’s performance across all scenarios. Such models are valuable for simulating EV behavior and analyzing mixed-autonomy traffic dynamics in EV-integrated environments.

\end{abstract}


\section{Introduction}

Electric vehicles (EVs) are becoming increasingly common on roads around the world. As EV adoption grows, it becomes essential to understand how these vehicles behave in mixed traffic environments, particularly in car-following situations \cite{matcha2020simulation}. Car-following behavior plays a key role in road safety, traffic flow, and the development of advanced driver assistance systems \cite{2.zhou2020modeling}. Over the years, many car-following models have been developed to simulate and analyze driver behavior. Traditional physics-based models such as the Intelligent Driver Model (IDM) \cite{Treiber2013IDM}, Optimal Velocity Model (OVM) \cite{Bando1995OVM}, and their extensions \cite{extension1, extension2, extension3} offer interpretable frameworks based on physical and behavioral principles. These models allow for analytical insights into vehicle interactions and are often used for ensuring safety and stability in simulations \cite{3.chen2023follownet}.

The IDM describes acceleration using factors like desired velocity, spacing, and time headway. Similarly, the OVM and its relative velocity variant (OVRV) compute target speeds based on headway and speed differences \cite{bando1995dynamical, jiang2001full}. While these models are simple and interpretable, they may not fully capture the complexity and variability found in real-world driving scenarios. To address these limitations, machine learning (ML) models have gained popularity due to their ability to learn directly from data. Supervised learning techniques such as decision trees, random forests, and neural networks can predict acceleration or spacing based on features like speed, relative position, and gap type \cite{ma2015lstm, liang2018deep}. ML models are particularly effective at modeling diverse driving behaviors, but they often lack physical interpretability and require large datasets for training \cite{matcha2020simulation} \cite{ 3.chen2023follownet}.

Recent research has explored hybrid approaches that integrate classical models with ML-based corrections. For example, residual learning frameworks use classical model predictions as a baseline and refine them using machine learning techniques \cite{chen2021residual}. These approaches aim to combine the strengths of both interpretability and data-driven flexibility. In the context of connected and automated vehicles, Cooperative Adaptive Cruise Control (CACC) models have drawn attention for their ability to enhance platoon stability using vehicle-to-vehicle (V2V) communication \cite{rajamani2011vehicle}. Simplified versions of CACC can still be used in the absence of connectivity, relying on local measurements such as spacing and relative velocity. Empirical studies have shown that CACC can reduce traffic disturbances and improve flow \cite{Milanes2014CACC}.

Efforts like FollowNet have introduced benchmarking datasets that compare classical, ML-based, and hybrid car-following models under realistic driving conditions \cite{Zhou2023FollowNet}. These studies help guide researchers in selecting models based on performance, interpretability, and adaptability to real-world variability. The field of car-following modeling has witnessed significant progress with the integration of advanced machine-learning techniques and hybrid approaches. For instance, the study in \cite{han2023ensemblefollower} introduced EnsembleFollower, a hierarchical framework that leverages reinforcement learning to dynamically select among multiple low-level car-following models, enhancing adaptability across diverse driving scenarios. In the realm of hybrid modeling, a recent study \cite{jiang2024generic} introduced a stochastic hybrid car-following model utilizing approximate Bayesian computation. This approach probabilistically integrates multiple car-following models, offering a flexible framework that captures the stochastic nature of driver behavior. 

An adaptive Kalman-based hybrid strategy was proposed \cite{zheng2023adaptive}, combining Twin Delayed Deep Deterministic Policy Gradient (TD3) with Cooperative Adaptive Cruise Control (CACC). This approach uses an adaptive coefficient derived from multi-timestep predictions to improve safety and ride comfort in mixed-traffic environments. Another recent study \cite{matcha2020simulation} introduced a hybrid car-following model that integrates kinetic dynamics with deep learning networks. By combining traditional mathematical frameworks with Long Short-Term Memory (LSTM) networks, the model effectively captures both general traffic patterns and individual driver characteristics, enhancing prediction accuracy and interpretability. These recent contributions underscore the ongoing evolution of car-following models, emphasizing the importance of hybrid approaches that blend physics-based insights with data-driven adaptability. Incorporating such models can lead to more robust and accurate representations of vehicle-following behavior, particularly in the context of electric vehicles operating in mixed-traffic environments.

This paper builds on these foundations by presenting a comparative analysis of classical and data-driven models for EV car-following behavior. We use a real-world dataset where an EV follows a conventional ICE vehicle under various gap settings. Classical models are calibrated by minimizing RMSE to observed data, while a Random Forest model is trained to predict acceleration based on spacing, speed, and gap type. The main objectives of this study are as follows:
\begin{itemize}
    \item We use real-world EV experiment data to calibrate classical car-following models and find the best parameters for each model.
    % \item We evaluate how well classical car-following models represent real-world EV-following behavior.
    % \item We demonstrate the potential of a data-driven machine learning approach (Random Forest) to capture EV-specific car-following behavior.
    \item We use these best parameters to simulate the classical models, and we also train a machine learning model to compare its performance with the classical models.
    \item Our study provides a comprehensive set of calibrated parameters offering valuable insights for future traffic simulation studies.
\end{itemize}

% The main objective is to assess each model’s predictive performance and to explore different classical and data-driven models on real-world experimental data. Our findings aim to support researchers and engineers in choosing the right model depending on their application, whether they require interpretability, accuracy, or both.

\section{Dataset}
% To model the dynamic traffic behavior of EVs in a microscopic environment, we utilized real-world experimental trajectory data, which is commercially available for EV-ACC vehicles. We used this leader-follower trajectory dataset to calibrate the classical models as well as the Random Forest Regressor to predict the acceleration. The dataset was collected as part of a comprehensive data collection campaign by Kan \cite{kan2021field}. The experiment was conducted in a controlled environment, where the ICE vehicle acted as the lead vehicle (Toyota Camry, 2021), and it followed a predefined speed profile, and an EV equipped with ACC (Hyundai IONIQ 5, 2022) followed the leader's trajectory.

To model the dynamic traffic behavior of EVs in a microscopic environment, we utilized real-world experimental trajectory data, which is commercially available for EV-ACC vehicles. The dataset was collected by Kan \cite{chon2021field}. The experiment was conducted in a controlled environment, where the ICE vehicle acted as the lead vehicle (Toyota Camry, 2021), and it followed a predefined speed profile, and an EV equipped with ACC (Hyundai IONIQ 5, 2022) followed the leader's trajectory. The positions of both the leader and follower vehicles were recorded using a Racebox GPS logger, which captures data at a frequency of 25 Hz and achieves a horizontal accuracy of approximately 10 cm. Vehicle spacing was then derived by applying the Haversine distance formula to the recorded positional data.

Both the leader and follower started with predefined initial spacing and acceleration. After reaching the free-flow speed, the subject vehicle activated the ACC and occasionally adjusted the acceleration above the designated point to maintain the spacing equilibrium. Once the leader vehicle reached the equilibrium, it decelerated to a congested speed and maintained that speed for the next 10 seconds, after that it returned to the previous free-flow speed. The leader vehicle's free-flow speed was set at four different levels for each gap configuration: 60 mph, 55 mph, 45 mph, and 35 mph. This sequence of speed fluctuation was repeated 8 times, with two repetitions for each of four gap settings provided by the vehicle's ACC system: short, medium, long, and extra-long, a total of 136 speed fluctuations in the final dataset.

% To prepare the data for modeling, several preprocessing steps were applied:

% \begin{itemize}
%     \item \textbf{Unit Conversion:} To maintain the consistency among the physics-based models, the Leader and follower vehicle speeds were transformed into meters per second (m/s) from kilometers per hour(km/h).
%     \item \textbf{Feature Engineering:} To describe the distance between follower and leader, a relative speed feature $\Delta v = v_{\text{leader}} - v_{\text{follower}}$ was computed.
%     \item \textbf{Encoding Categorical Features:} Categorical features(gap settings) were converted into one-hot encoded binary values to utilize in the ML model.
% \end{itemize}

\section{Classicial Models}
This section illustrates physics-based models used in the calibration and simulation using EV data.

\subsection{Intelligent Driver Model (IDM)}

The IDM is a widely adopted car-following model that computes acceleration based on spacing, speed, and desired time gap. The acceleration is given by:
\begin{equation}
a(t) = a_{\text{max}} \left[ 1 - \left( \frac{v(t)}{v_0} \right)^\delta - \left( \frac{s^*(t)}{s(t)} \right)^2 \right]
\end{equation}
where $v(t)$ is the current speed, $v_0$ is the desired speed, $\delta$ is the acceleration exponent and $s(t)$ is the actual spacing. The desired spacing $s^*(t)$ is defined as:
\begin{equation}
s^*(t) = s_0 + T \cdot v(t) + \frac{v(t) \cdot \Delta v(t)}{2 \sqrt{a_{\text{max}} b}}
\end{equation}

where $v_0$ is the desired speed, $T$ is the time headway, $s_0$ is the minimum spacing, $a_{\text{max}}$ is the maximum acceleration, and $b$ is the comfortable deceleration.

\subsection{Optimal Velocity Model (OVM)}

The OVM assumes that a driver adapts their acceleration based on the difference between the current speed and an optimal velocity determined by spacing:
\begin{equation}
a(t) = \alpha \left( v_{\text{opt}}(s) - v(t) \right)
\end{equation}
where $v_{\text{opt}}(s)$ is typically modeled as a sigmoid function:
\begin{equation}
v_{\text{opt}}(s) = v_{\text{max}} \cdot \tanh\left( \frac{s}{s_{st}} - 2 \right)
\end{equation}
where  $\alpha$ is sensitivity parameter, $v_{\text{max}}$ is the maximum velocity, and $s_{st}$ is the spacing scale factor.

\subsection{Optimal Velocity Relative Velocity Model (OVRV)}

The OVRV model extends OVM by incorporating the effect of the relative velocity between the lead and the following vehicles:
\begin{equation}
a(t) = k_1 \left( s(t) - \eta - \tau \cdot v(t) \right) + k_2 \left( v_l(t) - v(t) \right)
\end{equation}

Here, $k_1$ and $k_2$ are sensitivity parameters, $\eta$ is the spacing offset, $\tau$ is a delay-like factor, and $v_l(t)$ is the leader’s speed.

\subsection{Cooperative Adaptive Cruise Control (CACC)}

The CACC model incorporates cooperative behavior using a spacing control law and relative velocity. We use a simplified version that excludes leader acceleration:
\begin{equation}
a(t) = k_1 \left( v_l(t) - v(t) \right) + k_2 \left( s(t) - \left( s_0 + T \cdot v(t) \right) \right)
\end{equation}

Here, $k_1$ controls the response to speed difference, and $k_2$ scales the correction to spacing error. The desired spacing is modeled as $s_0 + T \cdot v(t)$, where $T$ is the time gap and $s_0$ is the minimum spacing. Their performance is evaluated in Section~\ref{Evaluation and Comparison}.

% Each of these models offers a different balance between simplicity, responsiveness, and realism. Their performance is evaluated in Section~\ref{Evaluation and Comparison}.

\section{Model Calibration}
In this section, we used real-world trajectory data of EVs to ensure the fair evaluation of classical car-following models.

\begin{table*}[!ht]
\centering
\caption{Best-calibrated parameters and RMSE across Gap Settings}
\begin{tabular}{l|c|c|c|c|c|c}
\hline
\multirow{2}{*}{\textbf{Models}} & \multicolumn{2}{c|}{\textbf{Medium Gap}} & \multicolumn{2}{c|}{\textbf{Long Gap}} & \multicolumn{2}{c}{\textbf{XLong Gap}} \\
\cline{2-7}
& \textbf{Parameters} & \textbf{RMSE} & \textbf{Parameters} & \textbf{RMSE} & \textbf{Parameters} & \textbf{RMSE} \\
\hline
IDM & $[22.12, 1.29, 1.73, 2.00, 2.00]$ & 93.79 & $[20.00, 1.29, 3.81, 2.00, 2.00]$ & 207.90 & $[20.15, 1.41, 4.71, 2.00, 1.11]$ & 237.81 \\
OVRV & $[0.34, 0.39, 9.92, 0.99]$ & 5.13 & $[0.60, 0.00, 8.40, 1.36]$ & 4.06 & $[0.69, 0.00, 12.54, 1.69]$ & 4.18 \\
OVM & $[21.88, 0.82, 9.43]$ & 53.69 & $[21.69, 1.49, 10.63]$ & 63.02 & $[21.78, 0.54, 12.73]$ & 68.47 \\
CACC & $[0.10, 0.39, 4.96, 1.47]$ & 3.06 & $[0.31, 0.28, 8.89, 1.44]$ & 2.67 & $[0.19, 0.28, 7.43, 2.00]$ & 2.88 \\
RF Regressor & -- & \textbf{0.0046} & -- & \textbf{0.0016} & -- & \textbf{0.0025} \\
\hline
\end{tabular}
\label{tab:calib_all}
\end{table*}

\subsection{Calibration Objective}

We calculated the Root Mean Squared Error (RMSE) between the simulated and experimental spacing as the calibration objective function using the following equation:
\begin{equation}
\text{RMSE} = \sqrt{\frac{1}{N} \sum_{i=1}^N (s_{\text{sim}}(i) - s_{\text{exp}}(i))^2}
\label{eq:rmse}
\end{equation}
where \(s_{\text{sim}}\) and \(s_{\text{exp}}\) represent the simulated and experimentally observed spacing at the $i$-th time step, respectively, and $N$ is the total number of observations.

\subsection{Optimization Strategy}

Each model was calibrated using the L-BFGS-B algorithm provided by \texttt{\textbf{scipy.optimize.minimize}}, which supports parameter box constraints. The dataset was sliced into six subsets to train the model, and calibration was performed separately on these slices (200s subsets) for each gap, and the best parameter set was selected based on overall RMSE performance. The subset with the lowest RMSE was selected for the final simulation. This ensured that the models were tested across a range of traffic conditions.

% \subsection{Parameter Settings}

% \begin{itemize}
%     \item \textbf{IDM}: Parameters optimized were \([v_0, T, s_0, a_{\text{max}}, b]\)
%     \item \textbf{OVM}: Parameters included \([v_{\text{max}}, \alpha, s_{st}]\)
%     \item \textbf{OVRV}: Parameters included \([k_1, k_2, \tau, \eta]\)
%     \item \textbf{CACC}: Parameters included \([k_1, k_2, k_3, s_0, T]\)
% \end{itemize}

% \subsection{Calibration Results}

\section{Machine Learning Models}
In addition to classical models, we trained data-driven ML models to predict vehicle behavior. Two separate models were developed: one for acceleration prediction and another for spacing prediction. Both were implemented using Random Forest Regressors. We selected the Random Forest regressor as our sole ML model. This choice stems from our aim to benchmark ML against classical methods, rather than exploring inter-ML comparisons. RF offers a robust, fast, and interpretable model that has proven superior to both traditional and neural methods in prior car-following studies \cite{shi2021data, choi2021comparison, rowan2025systematic}.

\subsection{Acceleration Prediction Model}
The goal of the first model is to predict the acceleration of the follower vehicle directly from input features such as spacing and speed. We consider follower speed, gap settings, speed fluctuation, and the speed difference $\Delta v$ between the follower and the lead vehicle.
We computed the target acceleration $a(t)$ using equation \ref{eq:custom_acc},
\begin{equation}
a(t) = \frac{v(t) - v(t-\Delta t)}{\Delta t}
\label{eq:custom_acc}
\end{equation}
It achieved high precision with RMSE around 0.68 m/s\textsuperscript{2} and $R^2$ close to 0.996 on the entire dataset.
% \begin{equation}
% a_{\text{pred}} = f(\text{spacing},\ v_{\text{follower}},\ \Delta v,\ \text{gap\_setting},\ \text{fluctuation})
% \end{equation}

% where $\Delta v = v_{\text{leader}} - v_{\text{follower}}$ is the relative speed, and $f(\cdot)$ is the regression function learned from data.

% The model was trained on a subset of the dataset and evaluated using RMSE and $R^2$. It achieved high precision with RMSE around 0.68 m/s\textsuperscript{2} and $R^2$ close to 0.996 on the entire dataset.

\subsection{Space Prediction Model}

The second ML model predicts the spacing between vehicles. The model was trained using the same dataset and evaluated against actual spacing measurements. The results showed a strong alignment across various time windows. \Remark{Can this be quantified?}

% \begin{equation}
% s_{\text{pred}} = g(v_{\text{follower}},\ v_{\text{leader}},\ \Delta v,\ \text{gap\_setting})
% \end{equation}

% \subsection{Evaluation and Integration}

% Both models were evaluated using residual plots, predicted versus actual graphs, and integration techniques. For example, predicted acceleration was integrated over time to estimate speed:

% \begin{equation}
% \hat{v}(t) = v(t-1) + a_{\text{pred}}(t-1) \cdot \Delta t
% \end{equation}

% Similarly, spacing predictions were compared across full sequences of data.

% While ML models demonstrated superior accuracy, they lack physical interpretability. They are best used when accuracy is critical and sufficient training data is available, while physics-based models remain preferable in interpretable and safety-critical applications.

\section{Evaluation and Comparison}
\label{Evaluation and Comparison}

In this section, we evaluate the performance of both classical and machine learning models by comparing their ability to reproduce real-world EV car-following behavior. The primary metric used to assess accuracy is the Root Mean Squared Error (RMSE) as shown in Equation~\eqref{eq:rmse}. For machine learning models, we also report the coefficient of determination ($R^2$), which measures how well the predictions explain the variance in the data.

% \subsection{Evaluation Metrics}

% \begin{equation}
% \text{RMSE} = \sqrt{\frac{1}{N} \sum_{i=1}^{N} (y_i - \hat{y}_i)^2}
% \end{equation}

% \begin{equation}
% R^2 = 1 - \frac{\sum (y_i - \hat{y}_i)^2}{\sum (y_i - \bar{y})^2}
% \end{equation}

% \begin{figure*}[t]
%     \centering
%     \begin{minipage}[b]{0.48\linewidth}
%         \centering
%         \includegraphics[width=\linewidth]{updated_rmse_medium.pdf}
%         \subcaption{}
%     \end{minipage}
%     \hfill
%     \begin{minipage}[b]{0.48\linewidth}
%         \centering
%         \includegraphics[width=\linewidth]{updated_rmse_medium.pdf}
%         \subcaption{}
%     \end{minipage}
    
%     \vspace{2mm} % vertical spacing between rows

%     \begin{minipage}[b]{0.5\linewidth}
%         \centering
%         \includegraphics[width=\linewidth]{updated_rmse_xlong.pdf}
%         \subcaption{}
%     \end{minipage}
    
%     \caption{RMSE comparison of CACC, OVRV, OVM, and IDM models across three gap settings: (a) Medium Gap — CACC and OVRV show low errors, while OVM and IDM have higher RMSE; (b) Long Gap — similar trend, with IDM showing substantial error; (c) Extra-Long Gap — IDM reaches the highest RMSE.}

%     \label{fig:three-image-layout}
% \end{figure*}
\begin{figure*}[t]
    \centering
    \begin{minipage}[b]{0.32\linewidth}
        \centering
        \includegraphics[width=\linewidth]{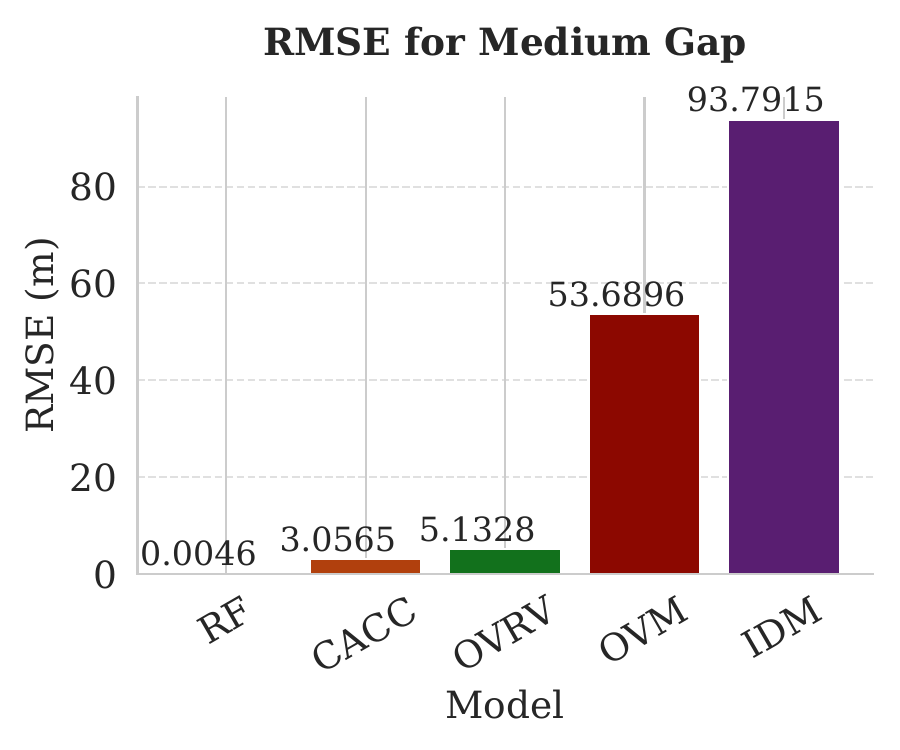}
        \subcaption{Medium Gap}
    \end{minipage}
    \hfill
    \begin{minipage}[b]{0.32\linewidth}
        \centering
        \includegraphics[width=\linewidth]{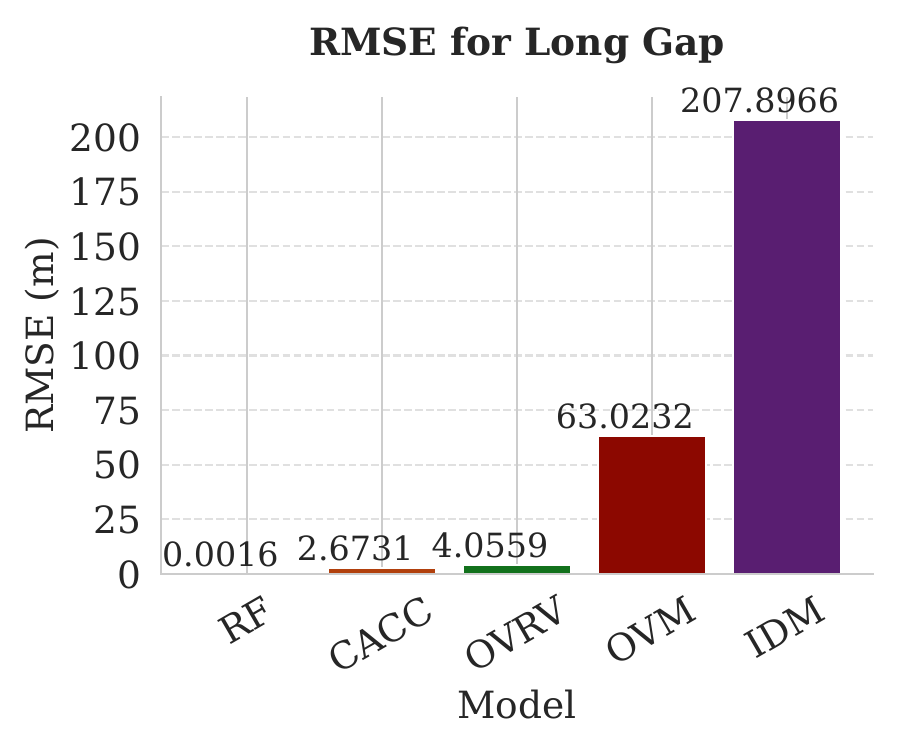}
        \subcaption{Long Gap}
    \end{minipage}
    \hfill
    \begin{minipage}[b]{0.32\linewidth}
        \centering
        \includegraphics[width=\linewidth]{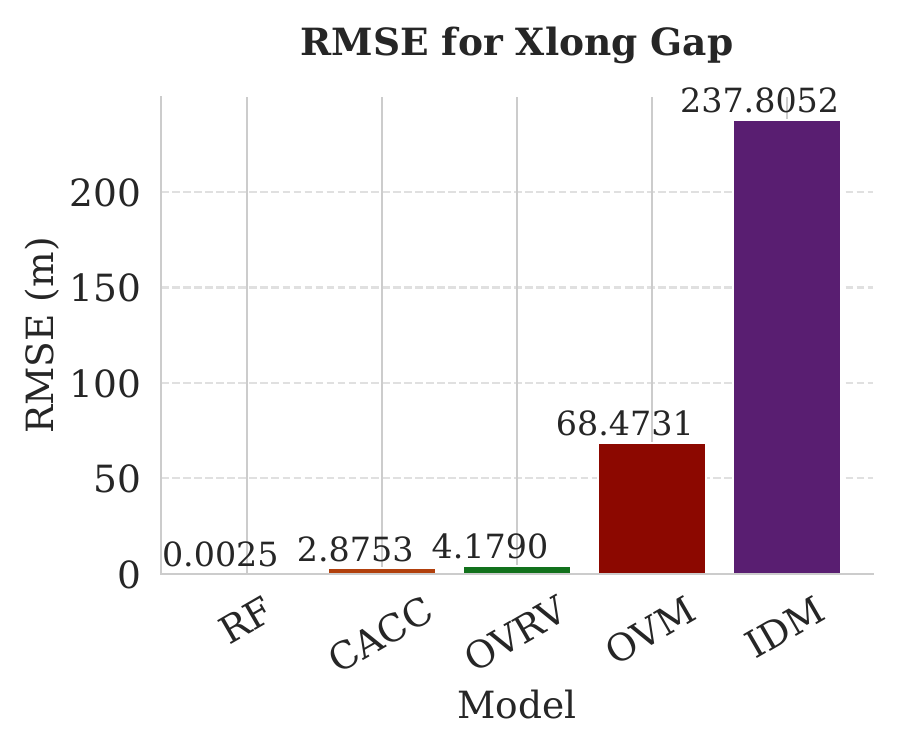}
        \subcaption{Extra-Long Gap}
    \end{minipage}
    
    \caption{RMSE comparison of RF, CACC, OVRV, OVM, and IDM models across three gap settings: (a) Medium Gap — RF, CACC and OVRV show low errors, while OVM and IDM have higher RMSE; (b) Long Gap — similar trend, with IDM showing substantial error; (c) Extra-Long Gap — IDM reaches the highest RMSE.}
    \label{fig:three-image-layout}
\end{figure*}

\subsection{Residual Analysis of Machine Learning Models}

Beyond numerical evaluation metrics, we also analyze the residuals (the difference between actual and predicted values) for both the acceleration and spacing prediction models. A residual plot is a common diagnostic tool used to assess model performance, revealing potential biases and non-random prediction patterns \cite{book1_chatterjee2015regression, book2_montgomery2021introduction}.

\begin{figure*}[t]
\centering
\begin{minipage}[b]{0.45\linewidth}
    \centering
    \includegraphics[width=\linewidth]{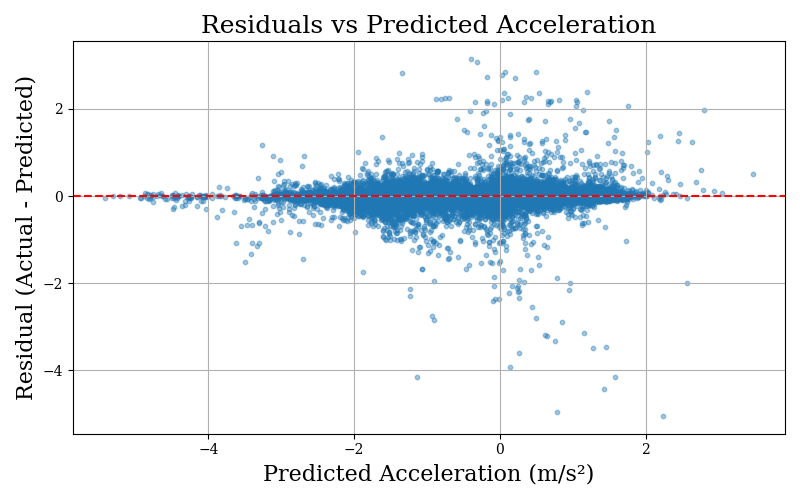}
    \subcaption{}
\end{minipage}
\begin{minipage}[b]{0.45\linewidth}
    \centering
    \includegraphics[width=\linewidth]{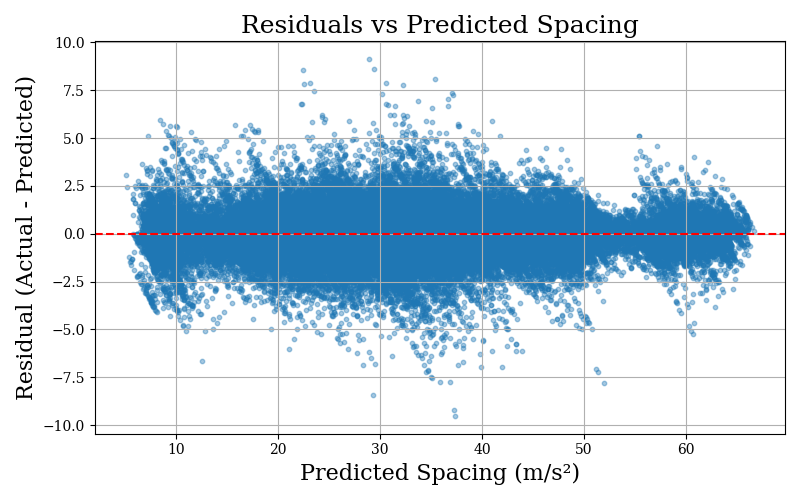}
    \subcaption{}
\end{minipage}
\caption{Most of the residual points are closely centered around $y=0$ line for both (a) and (b), which suggests that the models are not biased and do not overpredict or underpredict consistently. Both  (a) and (b) show slight deviation for very high and low data points, but effectively capture the complex dynamics.}
\label{fig:residual_plot}
    
\end{figure*}

Figure~\ref{fig:residual_plot}(a) shows the residuals of the acceleration model plotted against predicted acceleration. The points are tightly clustered around the zero-error line, indicating that the model captures the underlying trend well and does not exhibit systematic errors across different predicted values.

Similarly, Figure~\ref{fig:residual_plot}(b) presents the residuals for the spacing prediction model. Although there is a slightly wider spread compared to the acceleration model, the residuals remain largely centered around zero, suggesting that the model generalizes well over a wide range of spacing conditions.

\subsection{Models Performance Comparison}

Table~\ref{tab:calib_all} illustrates the best-calibrated parameters and RMSE for three gap settings. The parameters for each classical model are as follows: IDM - $(v_0, \tau, s_0, a, b)$, OVRV - $(k_1, k_2, \tau, \eta)$, OVM - $(v_{\max}, \alpha, s_{st})$, CACC - $(k_1, k_2, s_0, T)$
The capacity of classical physics-based models and ML models to replicate actual spacing and accelerating behavior is used to evaluate their performance.

The CACC consistently has better performance in all three gap settings than OVRV, OVM, and IDM. This is because the improved version has leader acceleration and spacing error feedback, which allows it to adjust the dynamic changes. OVRV also demonstrates commendable performance, especially in medium and long-gap settings. The performance of OVM and IDM was not as accurate as others. IDM overestimates spacing in larger gaps, and OVM much more deviates from the experimental one for spacing prediction.

The Random Forest regressor is trained to predict the acceleration and the RMSE of this model is as low as \textbf{0.0016} m for the long gap setting, which is much lower than all classical models in every gap category.  This illustrates how, without depending on predetermined physical assumptions, the model can directly extract complex nonlinear connections from the data.  The ML model's effectiveness is further supported by residual analysis, which indicates that errors are symmetrically distributed, centered around zero, and not the subject of significant bias or variance drift.

These findings imply that machine learning models, when given enough data, offer better accuracy and generalization than traditional models, which provide interpretation, a theoretical foundation, and more control over the assumptions.

\subsection{Visual Comparison}

\begin{figure*}[!ht]
    \centering
    \includegraphics[width=0.48\textwidth]{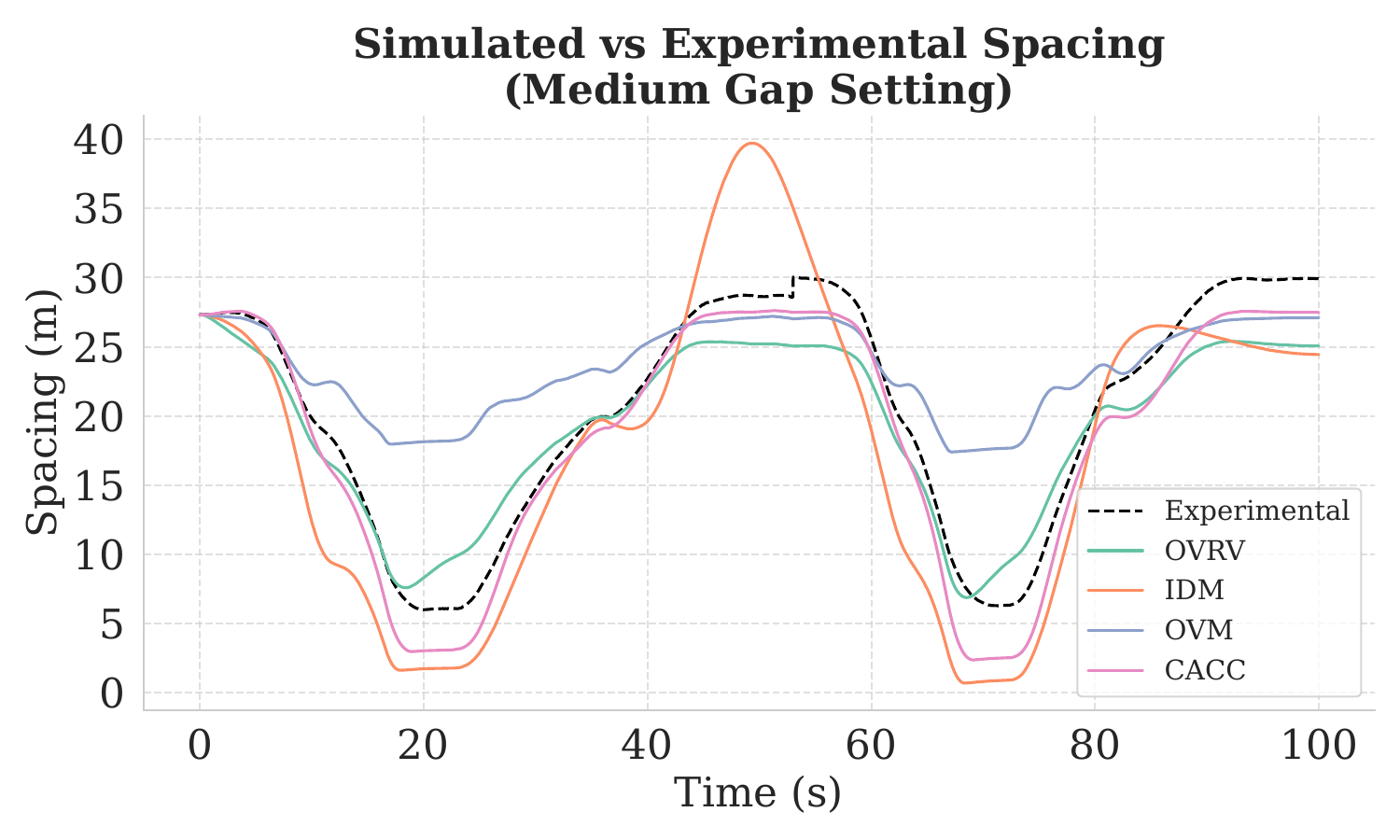}
    \includegraphics[width=0.48\textwidth]{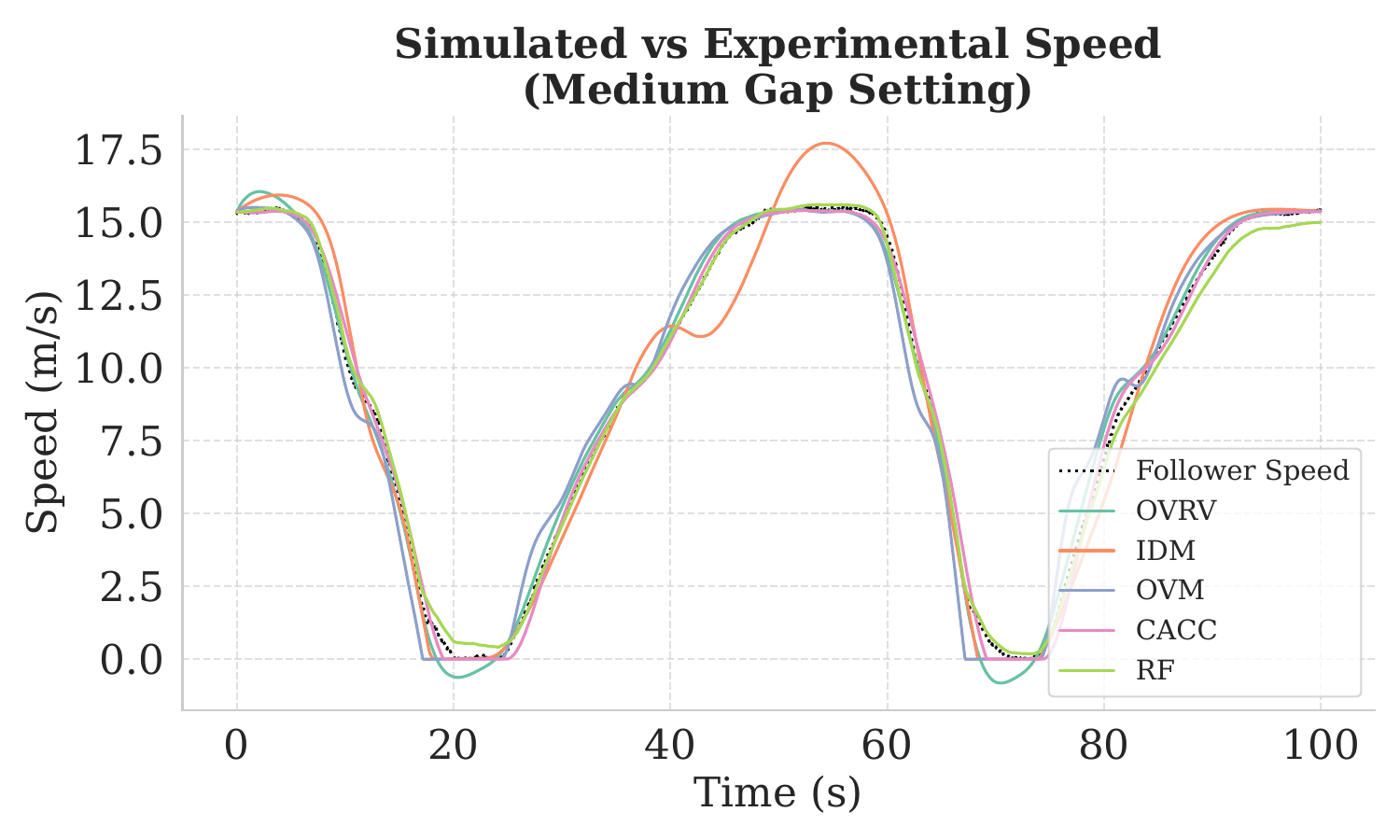}
    \caption{Simulated vs Experimental Spacing (left) and Speed (right) for Medium Gap}
    \label{fig:medium_gap_visuals}
\end{figure*}

\begin{figure*}[!ht]
    \centering
    \includegraphics[width=0.48\textwidth]{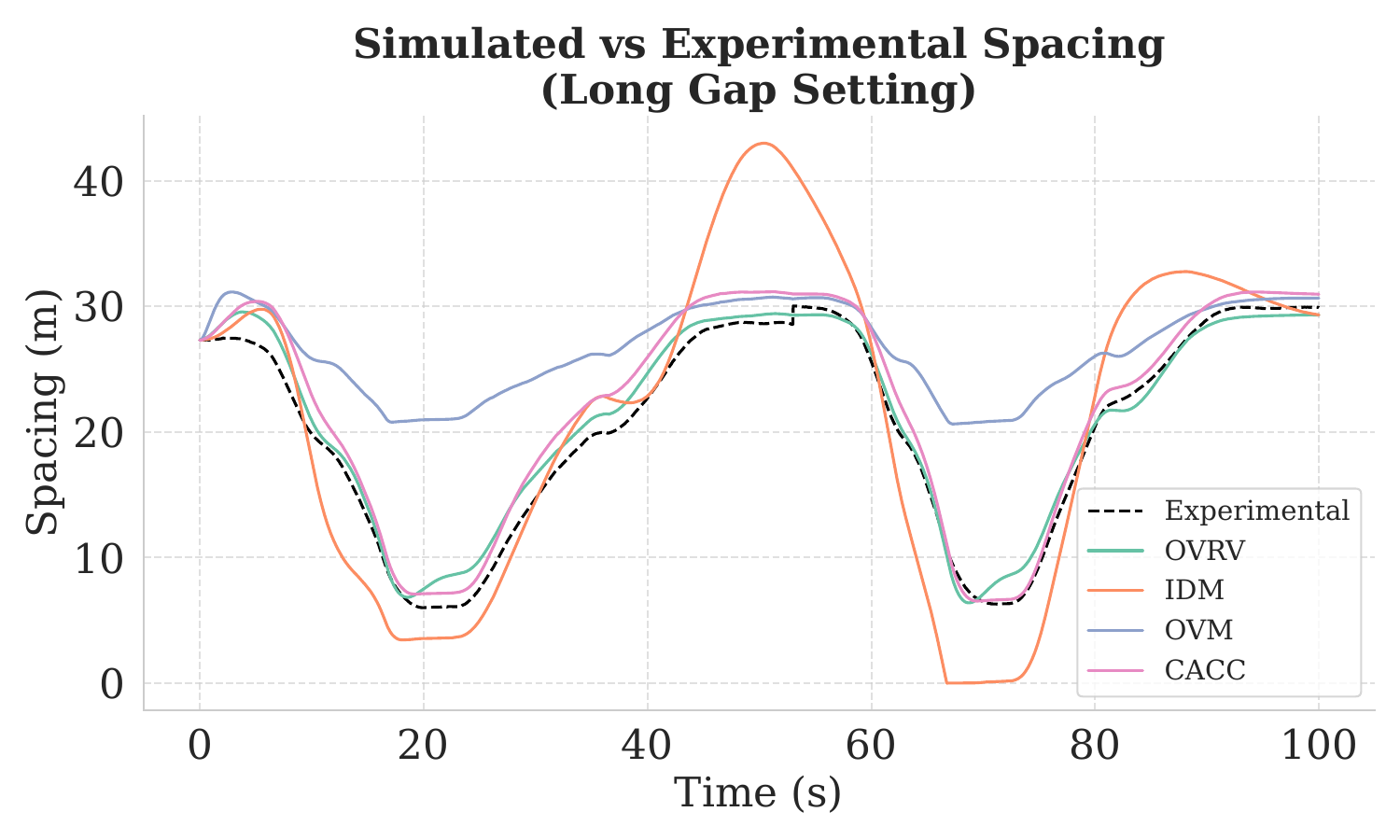}
    \includegraphics[width=0.48\textwidth]{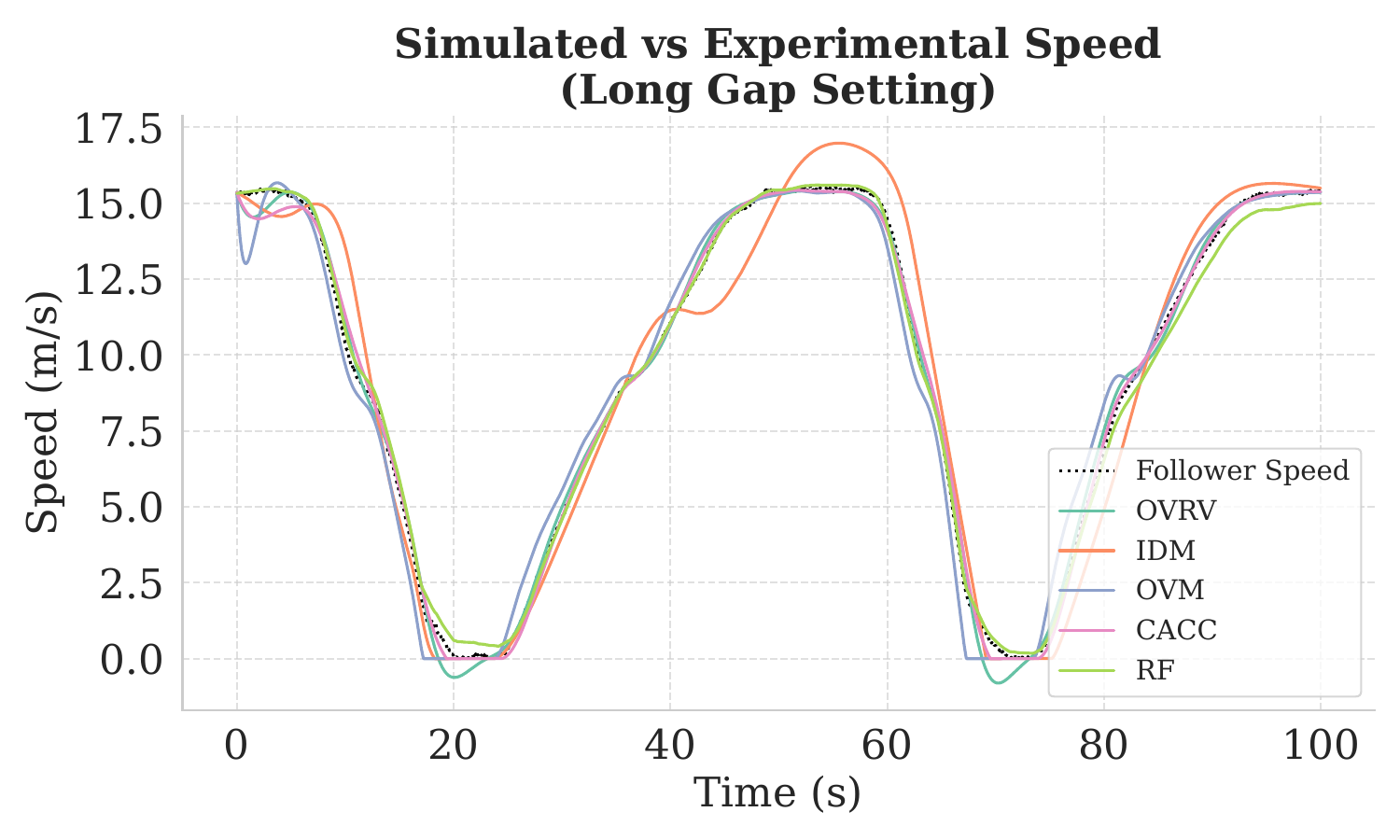}
    \caption{Simulated vs Experimental Spacing (left) and Speed (right) for Long Gap}
    \label{fig:long_gap_visuals}
\end{figure*}

\begin{figure*}[!ht]
    \centering
    \includegraphics[width=0.48\textwidth]{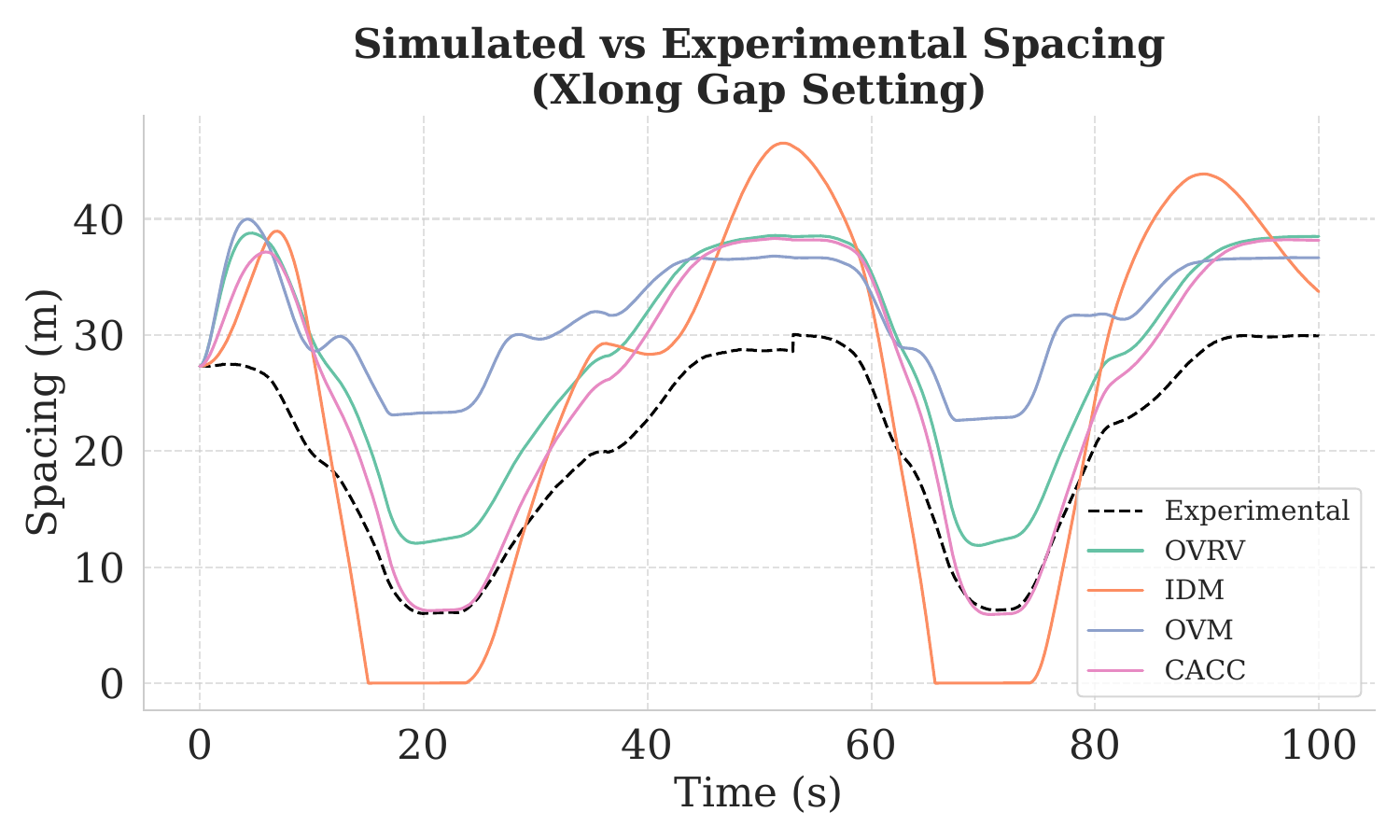}
    \includegraphics[width=0.48\textwidth]{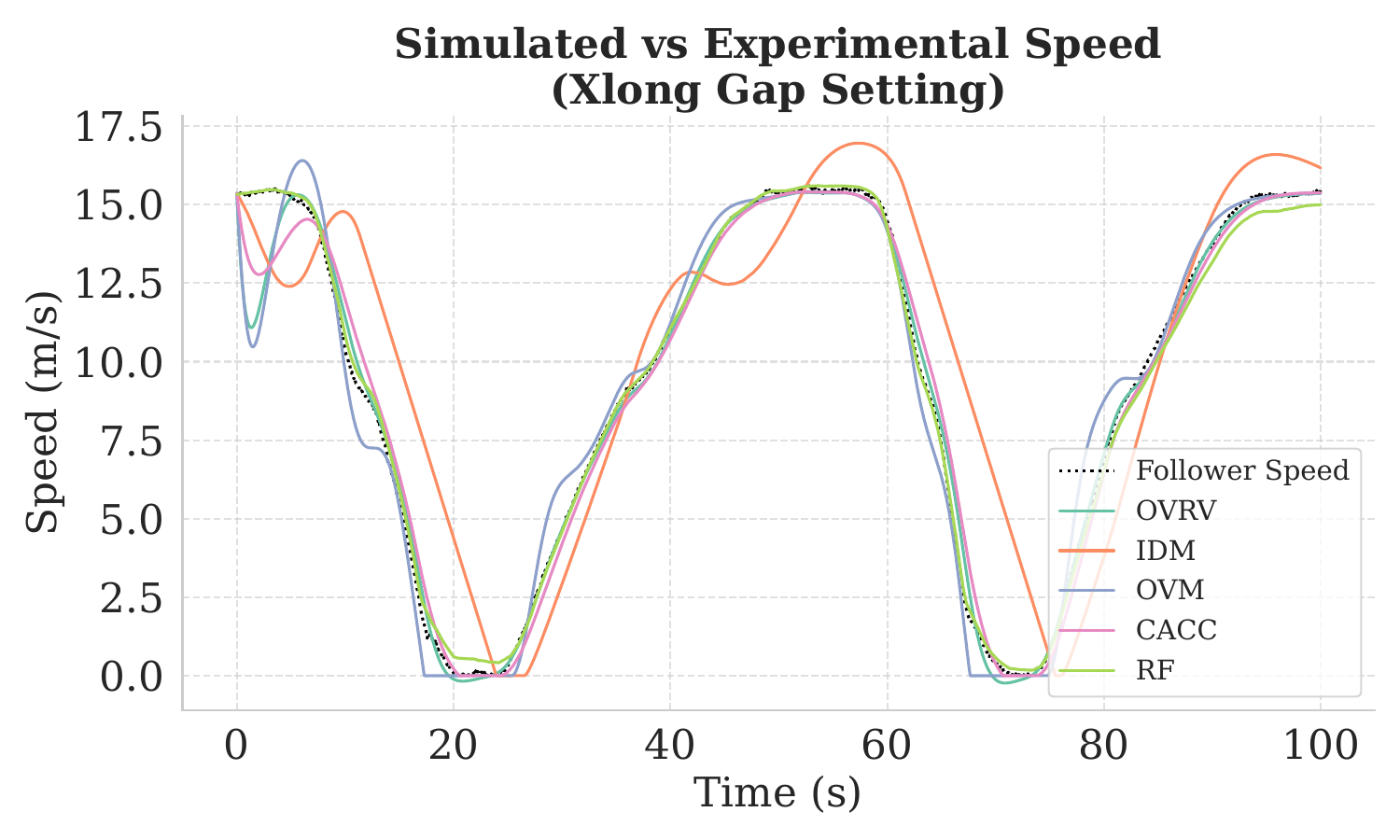}
    \caption{Simulated vs Experimental Spacing (left) and Speed (right) for XLong Gap}
    \label{fig:xlong_gap_visuals}
\end{figure*}

The Random Forest (RF) regressor demonstrates superior performance in mimicking the leader’s speed, particularly under the long gap setting, where it closely follows the actual trajectory. In contrast, for medium and extra long (xlong) gap settings, the RF model tends to be more conservative, slightly underestimating acceleration changes but still maintaining smooth transitions.

A remarkable pattern is observed during the initial 10 seconds across all gap settings -- classical physics-based models (IDM, OVM, OVRV, and CACC) exhibit greater fluctuations and lag in adapting to the leader's speed. On the other hand, the ML-based model shows immediate responsiveness and stability, indicating better generalization and adaptability from the start of the sequence. These results highlight the ML model's advantage in consistency and early accuracy, especially under complex or dynamically changing driving conditions.

Among physics-based models, CACC and OVRV align most closely with the leader’s speed. The CACC model notably maintains smooth acceleration profiles despite fluctuations in the leader’s velocity—a critical trait for reliable car-following behavior. At 38 and 85 seconds, where other models overreact or destabilize, CACC achieves smoother transitions without abrupt adjustments.

This stability is further evident in spacing plots (Figure~\ref{fig:medium_gap_visuals}, Figure~\ref{fig:long_gap_visuals} and Figure~\ref{fig:xlong_gap_visuals}), where CACC consistently preserves uniform inter-vehicle gaps. Its performance improves with larger gap settings which is suitable for real-world cooperative adaptive cruise control systems.

\section{Conclusion and Future Work}
% The results show a clear trade-off. Classical models are interpretable and grounded in physics, making them suitable for theoretical analysis and safety-critical systems. Machine learning models, though less transparent, deliver higher accuracy and adaptiveness, especially when trained on large and diverse datasets.
This paper presented a comparative study of classical and machine learning approaches to model the behavior of EV car-following scenarios. We used a real-world dataset in which an electric vehicle followed an ICE vehicle under various driving conditions.

We implemented and calibrated four classical models: the Intelligent Driver Model (IDM), Optimal Velocity Model (OVM), Optimal Velocity Relative Velocity (OVRV), and a simplified Cooperative Adaptive Cruise Control (CACC) model. These models were evaluated using RMSE against observed spacing and speed data. We also developed separate random forest regressors, one to predict acceleration and the other one to predict spacing. These models achieved high accuracy, outperforming classical models in accuracy metrics such as RMSE.

While classical models offer simplicity, interpretability, and robustness, machine learning models provide better accuracy and adaptability to varying driving patterns. However, ML models may suffer from a lack of explainability, which can be critical in safety-sensitive applications.

% As a future direction, we propose developing hybrid models that combine classical physics-based models with data-driven machine-learning techniques. One promising approach is residual learning, where an ML model is trained to predict the error or correction to a baseline physics-based model.
Future work will focus on assessing the practical applicability of these models in real-world scenarios, we propose evaluating the computational efficiency and inference time of each model. This will provide insights into their feasibility for real-time deployment in electric vehicles for automation. Another important extension is to broaden the analysis to multi-vehicle scenarios or vehicle platoons, which will allow for a deeper understanding of the models’ scalability, interaction dynamics, and collective stability in more complex traffic environments.

\textbf{Code and Data Availability.} Code and Data is available on \url{https://github.com/AARC-lab/EV_Carfollowing_Calibration_ITSC25}.

% Such hybrid frameworks can leverage the strengths of both modeling paradigms to build safer, smarter, and more reliable vehicle follow-up systems in mixed-traffic environments.

{
\small 
\bibliographystyle{IEEEtran}
\bibliography{main}
}

\end{document}